\documentclass{article}
\pdfoutput=1
\usepackage[nonatbib,final]{nips_2016}

\usepackage[utf8]{inputenc} 
\usepackage[T1]{fontenc}    
\usepackage{hyperref}       
\usepackage{booktabs}       
\usepackage{amsfonts}       
\usepackage{nicefrac}       
\usepackage{graphicx}       
\usepackage{microtype}      
\usepackage[autostyle]{csquotes}
\usepackage{amsmath}
\usepackage{multirow}

\usepackage{booktabs}       

\title{Interaction Networks for Learning about Objects, Relations and
  Physics}

\author{
  Peter W.~Battaglia \\
  Google DeepMind\\
  London, UK N1C 4AG \\
  \texttt{peterbattaglia@google.com} \\
  \And
  Razvan~Pascanu \\
  Google DeepMind\\
  London, UK N1C 4AG \\
  \texttt{razp@google.com} \\
  \And
  Matthew~Lai \\
  Google DeepMind\\
  London, UK N1C 4AG \\
  \texttt{matthewlai@google.com} \\
  \And
  Danilo~Rezende \\
  Google DeepMind\\
  London, UK N1C 4AG \\
  \texttt{danilor@google.com} \\
  \And
  Koray~Kavukcuoglu \\
  Google DeepMind\\
  London, UK N1C 4AG \\
  \texttt{korayk@google.com} \\
}

\newcommand{\DS}{D_S}
\newcommand{\DR}{D_R}
\newcommand{\DX}{D_X}
\newcommand{\NO}{N_O}
\newcommand{\NR}{N_R}
\newcommand{\DE}{D_E}
\newcommand{\DP}{D_P}
\newcommand{\DA}{D_A}

\begin{document}

\maketitle

\begin{abstract}
  Reasoning about objects, relations, and physics is central to human
  intelligence, and a key goal of artificial intelligence. Here we
  introduce the \textit{interaction network}, a model which can reason
  about how objects in complex systems interact, supporting dynamical
  predictions, as well as inferences about the abstract properties of
  the system.  Our model takes graphs as input, performs object- and
  relation-centric reasoning in a way that is analogous to a
  simulation, and is implemented using deep neural networks. We
  evaluate its ability to reason about several challenging physical
  domains: n-body problems, rigid-body collision, and non-rigid
  dynamics. Our results show it can be trained to accurately simulate
  the physical trajectories of dozens of objects over thousands of
  time steps, estimate abstract quantities such as energy, and
  generalize automatically to systems with different numbers and
  configurations of objects and relations. Our interaction network
  implementation is the first general-purpose, learnable physics
  engine, and a powerful general framework for reasoning about object
  and relations in a wide variety of complex real-world domains.
\end{abstract}

\section{Introduction}
Representing and reasoning about objects, relations and physics is a
``core'' domain of human common sense knowledge \cite{Spelke1992}, and
among the most basic and important aspects of intelligence
\cite{tenenbaum2011grow,lake2016building}. Many everyday problems,
such as predicting what will happen next in physical environments or
inferring underlying properties of complex scenes, are challenging
because their elements can be composed in combinatorially many
possible arrangements. People can nevertheless solve such problems by
decomposing the scenario into distinct objects and relations, and
reasoning about the consequences of their interactions and
dynamics. Here we introduce the \textit{interaction network} -- a
model that can perform an analogous form of reasoning about objects
and relations in complex systems.

Interaction networks combine three powerful approaches: structured
models, simulation, and deep learning. Structured models
\cite{ghahramani2015probabilistic} can exploit rich, explicit
knowledge of relations among objects, independent of the objects
themselves, which supports general-purpose reasoning across diverse
contexts. Simulation is an effective method for approximating
dynamical systems, predicting how the elements in a complex system are
influenced by interactions with one another, and by the dynamics of
the system. Deep learning \cite{schmidhuber2015deep,lecun2015deep}
couples generic architectures with efficient optimization algorithms
to provide highly scalable learning and inference in challenging
real-world settings.

\begin{figure}[t]
  \centering
    \includegraphics[width=1\textwidth]{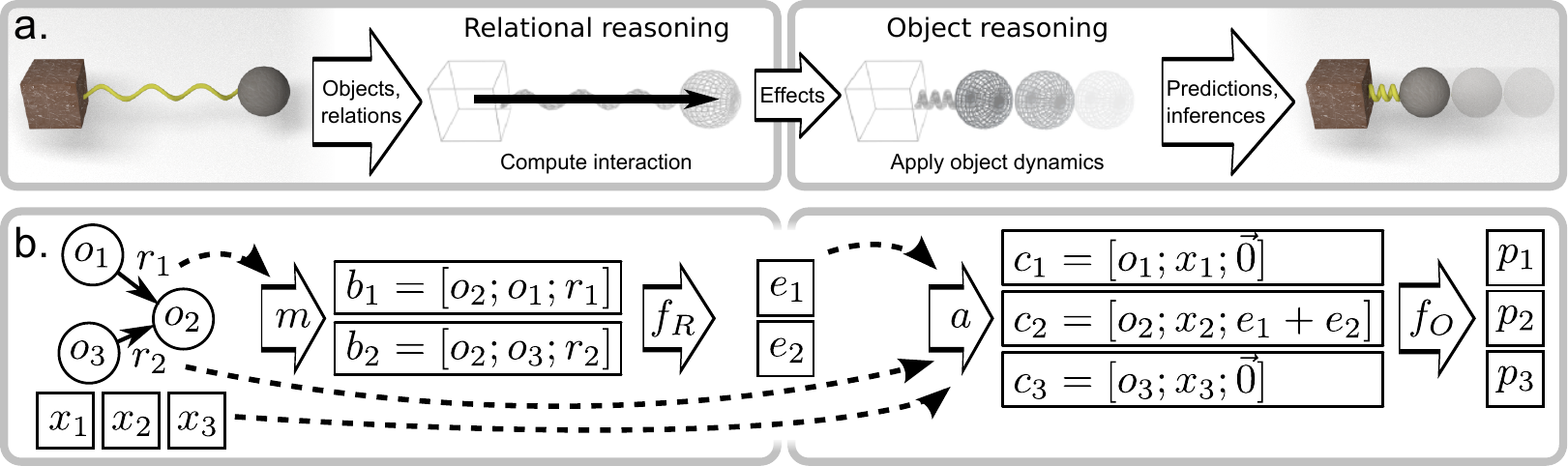}
    \caption{\small Schematic of an interaction network. \textit{a.}
      For physical reasoning, the model takes objects and relations as
      input, reasons about their interactions, and applies the effects
      and physical dynamics to predict new states. \textit{b.} For
      more complex systems, the model takes as input a graph that
      represents a system of objects, $o_j$, and relations, $\langle
      i,j,r_k \rangle_k$, instantiates the pairwise interaction terms,
      $b_k$, and computes their effects, $e_k$, via a relational
      model, $f_R(\cdot)$. The $e_k$ are then aggregated and combined
      with the $o_j$ and external effects, $x_j$, to generate input
      (as $c_j$), for an object model, $f_O(\cdot)$, which predicts
      how the interactions and dynamics influence the objects, $p$.}
    \label{fig:schematic}
\end{figure}

Interaction networks explicitly separate how they reason about
relations from how they reason about objects, assigning each task to
distinct models which are: fundamentally object- and relation-centric;
and independent of the observation modality and task specification
(see Model section 2 below and Fig.~\ref{fig:schematic}a). This lets
interaction networks automatically generalize their learning across
variable numbers of arbitrarily ordered objects and relations, and
also recompose their knowledge of entities and interactions in novel
and combinatorially many ways. They take relations as explicit input,
allowing them to selectively process different potential interactions
for different input data, rather than being forced to consider every
possible interaction or those imposed by a fixed architecture.

We evaluate interaction networks by testing their ability to make
predictions and inferences about various physical systems, including
n-body problems, and rigid-body collision, and non-rigid dynamics. Our
interaction networks learn to capture the complex interactions that
can be used to predict future states and abstract physical properties,
such as energy. We show that they can roll out thousands of realistic
future state predictions, even when trained only on single-step
predictions. We also explore how they generalize to novel systems with
different numbers and configurations of elements. Though they are not
restricted to physical reasoning, the interaction networks used here
represent the first general-purpose learnable physics engine, and even
have the potential to learn novel physical systems for which no
physics engines currently exist.

\paragraph{Related work}
Our model draws inspiration from previous work that reasons about
graphs and relations using neural networks. The ``graph neural
network'' \cite{scarselli09graphnet,li2015gated} is a framework that
shares learning across nodes and edges, the ``recursive autoencoder''
\cite{socher2011dynamic} adapts its processing architecture to exploit
an input parse tree, the ``neural programmer-interpreter''
\cite{reed2016npi} is a composable network that mimics the execution
trace of a program, and the ``spatial transformer''
\cite{jaderberg2015spatial} learns to dynamically modify network
connectivity to capture certain types of interactions. Others have
explored deep learning of logical and arithmetic relations
\cite{sutskever2009sym}, and relations suitable for visual
question-answering \cite{andreas2016qa}.

The behavior of our model is similar in spirit to a physical
simulation engine \cite{baraff2001physically}, which generates
sequences of states by repeatedly applying rules that approximate the
effects of physical interactions and dynamics on objects over
time. The interaction rules are relation-centric, operating on two or
more objects that are interacting, and the dynamics rules are
object-centric, operating on individual objects and the aggregated
effects of the interactions they participate in. A similar approach
has been developed independently by Chang et
al. \cite{chang2016compositional}.

Previous AI work on physical reasoning explored commonsense knowledge,
qualitative representations, and simulation techniques for
approximating physical prediction and inference
\cite{winston1975psychology,hayes1978naive}
. The
``NeuroAnimator'' \cite{grzeszczuk1998neuroanimator} was perhaps the
first quantitative approach to learning physical dynamics, by training
neural networks to predict and control the state of articulated
bodies.  Ladick\'{y} et al. \cite{jeong2015data} recently used
regression forests to learn fluid dynamics. Recent advances in
convolutional neural networks (CNNs) have led to efforts that learn to
predict coarse-grained physical dynamics from images
\cite{mottaghi2015newtonian,lerer2016learning,li2016fall}. Notably,
Fragkiadaki et al. \cite{fragkiadaki2015learning} used CNNs to predict
and control a moving ball from an image centered at its
coordinates. Mottaghi et al. \cite{mottaghi2016happens} trained CNNs
to predict the 3D trajectory of an object after an external impulse is
applied. Wu et al. \cite{wu2015galileo} used CNNs to parse objects
from images, which were then input to a physics engine that supported
prediction and inference.

\section{Model}
\paragraph{Definition}

To describe our model, we use physical reasoning as an example
(Fig.~\ref{fig:schematic}a), and build from a simple model to the full
interaction network (abbreviated IN). To predict the dynamics of a
single object, one might use an object-centric function, $f_O$, which
inputs the object's state, $o_t$, at time $t$, and outputs a future
state, $o_{t+1}$. If two or more objects are governed by the same
dynamics, $f_O$ could be applied to each, independently, to predict
their respective future states. But if the objects interact with one
another, then $f_O$ is insufficient because it does not capture their
relationship. Assuming two objects and one directed relationship,
e.g., a fixed object attached by a spring to a freely moving mass, the
first (the \textit{sender}, $o_1$) influences the second (the
\textit{receiver}, $o_2$) via their interaction. The effect of this
interaction, $e_{t+1}$, can be predicted by a relation-centric
function, $f_R$. The $f_R$ takes as input $o_1$, $o_2$, as well as
attributes of their relationship, $r$, e.g., the spring constant. The
$f_O$ is modified so it can input both $e_{t+1}$ and the receiver's
current state, $o_{2,t}$, enabling the interaction to influence its
future state, $o_{2,t+1}$,
\begin{equation*}
\begin{aligned}[c]
  e_{t+1} &= f_R(o_{1,t}, o_{2,t}, r)
\end{aligned}
\qquad \qquad
\begin{aligned}[c]
  o_{2,t+1} &= f_O(o_{2,t}, e_{t+1})
\end{aligned}
\end{equation*}

The above formulation can be expanded to larger and more complex
systems by representing them as a graph, $G = \langle O,R \rangle$,
where the nodes, $O$, correspond to the objects, and the edges, $R$,
to the relations (see Fig.~\ref{fig:schematic}b). We assume an
attributed, directed multigraph because the relations have attributes,
and there can be multiple distinct relations between two objects
(e.g., rigid and magnetic interactions). For a system with $\NO$
objects and $\NR$ relations, the inputs to the IN are,
\begin{equation*}
\begin{aligned}[t]
  O &= \{o_j\}_{j=1\dots\NO}
\end{aligned}\hspace*{-4pt},\;
\begin{aligned}[t]
  R &= \{\langle i, j, r_k \rangle_k\}_{k=1\dots\NR} \mbox{ where } i
  \ne j, 1 \le i,j \le \NO
\end{aligned}\hspace*{-4pt},\;
\begin{aligned}[t]
  X &= \{x_j\}_{j=1\dots\NO}
\end{aligned}
\end{equation*}
The $O$ represents the states of each object. The triplet, $\langle i,
j, r_k \rangle_k$, represents the $k$-th relation in the system, from
sender, $o_i$, to receiver, $o_j$, with relation attribute, $r_k$. The
$X$ represents external effects, such as active control inputs or
gravitational acceleration, which we define as not being part of the
system, and which are applied to each object separately.

The basic IN is defined as,
\begin{align}
  \mathrm{IN}(G) &= \phi_O(a(G, \; X, \; \phi_R(\; m(G)\;)\;))
\end{align}
\begin{equation}
\begin{aligned}
  & m(G) &= \: B \: &= \: \{b_k\}_{k=1\dots\NR} \\
  & f_R(b_k) &= \: e_k & \\
  & \phi_R(B) &= \: E \: &= \: \{e_k\}_{k=1\dots\NR}
\end{aligned}
\qquad \qquad \qquad
\begin{aligned}
  & a(G, X, E) &= \: C \: &= \: \{c_j\}_{j=1\dots\NO} \\
  & f_O(c_j) &= \: p_j & \\
  & \phi_O(C) &= \: P \: &= \: \{p_j\}_{j=1\dots\NO}
\end{aligned}
\end{equation}
The marshalling function, $m$, rearranges the objects and relations
into interaction terms, $b_k = \langle o_i, o_j, r_k \rangle \in B$,
one per relation, which correspond to each interaction's receiver,
sender, and relation attributes. The relational model, $\phi_R$,
predicts the effect of each interaction, $e_k \in E$, by applying
$f_R$ to each $b_k$. The aggregation function, $a$, collects all
effects, $e_k \in E$, that apply to each receiver object, merges them,
and combines them with $O$ and $X$ to form a set of object model
inputs, $c_j \in C$, one per object. The object model, $\phi_O$,
predicts how the interactions and dynamics influence the objects by
applying $f_O$ to each $c_j$, and returning the results, $p_j \in
P$. This basic IN can predict the evolution of states in a dynamical
system -- for physical simulation, $P$ may equal the future states of
the objects, $O_{t+1}$.

The IN can also be augmented with an additional component to make
abstract inferences about the system. The $p_j \in P$, rather than
serving as output, can be combined by another aggregation function,
$g$, and input to an abstraction model, $\phi_A$, which returns a
single output, $q$, for the whole system. We explore this variant in
our final experiments that use the IN to predict potential energy.

An IN applies the same $f_R$ and $f_O$ to every $b_k$
and $c_j$, respectively, which makes their relational and object
reasoning able to handle variable numbers of arbitrarily ordered
objects and relations. But one additional constraint must be satisfied
to maintain this: the $a$ function must be commutative and associative
over the objects and relations. Using summation within $a$ to merge
the elements of $E$ into $C$ satisfies this, but division would not.

Here we focus on binary relations, which means there is one
interaction term per relation, but another option is to have the
interactions correspond to $n$-th order relations by combining $n$
senders in each $b_k$. The interactions could even have variable
order, where each $b_k$ includes all sender objects that interact with
a receiver, but would require a $f_R$ than can handle variable-length
inputs. These possibilities are beyond the scope of this work, but are
interesting future directions.

\paragraph{Implementation}

The general definition of the IN in the previous section is agnostic
to the choice of functions and algorithms, but we now outline a
learnable implementation capable of reasoning about complex systems
with nonlinear relations and dynamics. We use standard deep neural
network building blocks, multilayer perceptrons (MLP), matrix
operations, etc., which can be trained efficiently from data using
gradient-based optimization, such as stochastic gradient descent.

We define $O$ as a $\DS \times \NO$ matrix, whose columns correspond
to the objects' $\DS$-length state vectors. The relations are a
triplet, $R=\langle R_r, R_s,R_a \rangle$, where $R_r$ and $R_s$ are
$\NO \times \NR$ binary matrices which index the receiver and sender
objects, respectively, and $R_a$ is a $\DR \times \NR$ matrix whose
$\DR$-length columns represent the $\NR$ relations' attributes. The
$j$-th column of $R_r$ is a one-hot vector which indicates the
receiver object's index; $R_s$ indicates the sender similarly. For the
graph in Fig.~\ref{fig:schematic}b, $R_r = \left[ \begin{smallmatrix}
    0 & 0 \\
    1 & 1 \\
    0 & 0
\end{smallmatrix}\right]$ and
$R_s = \left[ \begin{smallmatrix}
    1 & 0 \\
    0 & 0 \\
    0 & 1
\end{smallmatrix}\right]$.
The $X$ is a $\DX \times \NO$ matrix, whose columns are $\DX$-length
vectors that represent the external effect applied each of the $\NO$
objects.

The marshalling function, $m$, computes the matrix products, $O R_r$
and $O R_s$, and concatenates them with $R_a$: \quad  $ m(G) \; = \;  [ O R_r ; O R_s; R_a] \; = \; B$ . \\
The resulting $B$ is a $(2 \DS + \DR) \times \NR$ matrix, whose
columns represent the interaction terms, $b_k$, for the $\NR$
relations (we denote vertical and horizontal matrix concatenation with
a semicolon and comma, respectively). The way $m$ constructs
interaction terms can be modified, as described in our Experiments
section (3).

The $B$ is input to $\phi_R$, which applies $f_R$, an MLP, to each
column. The output of $f_R$ is a $\DE$-length vector, $e_k$, a
distributed representation of the effects. The $\phi_R$ concatenates
the $\NR$ effects to form the $\DE \times \NR$ effect matrix, $E$.

The $G$, $X$, and $E$ are input to $a$, which computes the $\DE \times
\NO$ matrix product, $\bar{E} = E R_r^T$, whose $j$-th column is
equivalent to the elementwise sum across all $e_k$ whose corresponding
relation has receiver object, $j$. The $\bar{E}$ is concatenated with
$O$ and $X$: \quad $a(G, X, E) \; = \; [ O ; X; \bar{E}] \; = \; C$. \\
The resulting $C$ is a $(\DS + \DX + \DE) \times \NO$ matrix, whose
$\NO$ columns represent the object states, external effects, and
per-object aggregate interaction effects.

The $C$ is input to $\phi_O$, which applies $f_O$, another MLP, to
each of the $\NO$ columns. The output of $f_O$ is a $\DP$-length
vector, $p_j$, and $\phi_O$ concatenates them to form the output
matrix, $P$.

To infer abstract properties of a system, an additional $\phi_A$ is
appended and takes $P$ as input. The $g$ aggregation function performs
an elementwise sum across the columns of $P$ to return a $\DP$-length
vector, $\bar{P}$. The $\bar{P}$ is input to $\phi_A$, another MLP,
which returns a $\DA$-length vector, $q$, that represents an abstract,
global property of the system.

Training an IN requires optimizing an objective function over the
learnable parameters of $\phi_R$ and $\phi_O$. Note, $m$ and $a$
involve matrix operations that do not contain learnable parameters.

Because $\phi_R$ and $\phi_O$ are shared across all relations and
objects, respectively, training them is statistically efficient. This
is similar to CNNs, which are very efficient due to their
weight-sharing scheme. A CNN treats a local neighborhood of pixels as
related, interacting entities: each pixel is effectively a receiver
object and its neighboring pixels are senders. The convolution
operator is analogous to $\phi_R$, where $f_R$ is the local
linear/nonlinear kernel applied to each neighborhood. Skip
connections, recently popularized by residual networks, are loosely
analogous to how the IN inputs $O$ to both $\phi_R$ and $\phi_O$,
though in CNNs relation- and object-centric reasoning are not
delineated. But because CNNs exploit local interactions in a fixed way
which is well-suited to the specific topology of images, capturing
longer-range dependencies requires either broad, insensitive
convolution kernels, or deep stacks of layers, in order to implement
sufficiently large receptive fields. The IN avoids this restriction by
being able to process arbitrary neighborhoods that are explicitly
specified by the $R$ input.

\section{Experiments}
\paragraph{Physical reasoning tasks}

Our experiments explored two types of physical reasoning tasks:
predicting future states of a system, and estimating their abstract
properties, specifically potential energy. We evaluated the IN's
ability to learn to make these judgments in three complex physical
domains: n-body systems; balls bouncing in a box; and strings composed
of springs that collide with rigid objects. We simulated the 2D
trajectories of the elements of these systems with a physics engine,
and recorded their sequences of states. See the Supplementary Material
for full details.

In the n-body domain, such as solar systems, all $n$ bodies exert
distance- and mass-dependent gravitational forces on each other, so
there were $n(n-1)$ relations input to our model. Across simulations,
the objects' masses varied, while all other fixed attributes were held
constant. The training scenes always included 6 bodies, and for
testing we used 3, 6, and 12 bodies. In half of the systems, bodies
were initialized with velocities that would cause stable orbits, if
not for the interactions with other objects; the other half had random
velocities.

In the bouncing balls domain, moving balls could collide with each
other and with static walls. The walls were represented as objects
whose shape attribute represented a rectangle, and whose inverse-mass
was 0. The relations input to the model were between the $n$ objects
(which included the walls), for ($n(n-1)$ relations). Collisions are
more difficult to simulate than gravitational forces, and the data
distribution was much more challenging: each ball participated in a
collision on less than 1\% of the steps, following straight-line
motion at all other times. The model thus had to learn that despite
there being a rigid relation between two objects, they only had
meaningful collision interactions when they were in contact. We also
varied more of the object attributes -- shape, scale and mass (as
before) -- as well as the coefficient of restitution, which was a
relation attribute. Training scenes contained 6 balls inside a box
with 4 variably sized walls, and test scenes contained either 3, 6, or
9 balls.

The string domain used two types of relations (indicated in $r_k$),
relation structures that were more sparse and specific than
all-to-all, as well as variable external effects. Each scene contained
a string, comprised of masses connected by springs, and a static,
rigid circle positioned below the string. The $n$ masses had spring
relations with their immediate neighbors ($2 (n-1)$), and all masses
had rigid relations with the rigid object ($2 n$). Gravitational
acceleration, with a magnitude that was varied across simulation runs,
was applied so that the string always fell, usually colliding with the
static object. The gravitational acceleration was an external input
(not to be confused with the gravitational attraction relations in the
n-body experiments). Each training scene contained a string with 15
point masses, and test scenes contained either 5, 15, or 30 mass
strings. In training, one of the point masses at the end of the
string, chosen at random, was always held static, as if pinned to the
wall, while the other masses were free to move. In the test
conditions, we also included strings that had both ends pinned, and no
ends pinned, to evaluate generalization.

\begin{figure}[t!]
  \centering
  \includegraphics[height=0.8324\textheight]{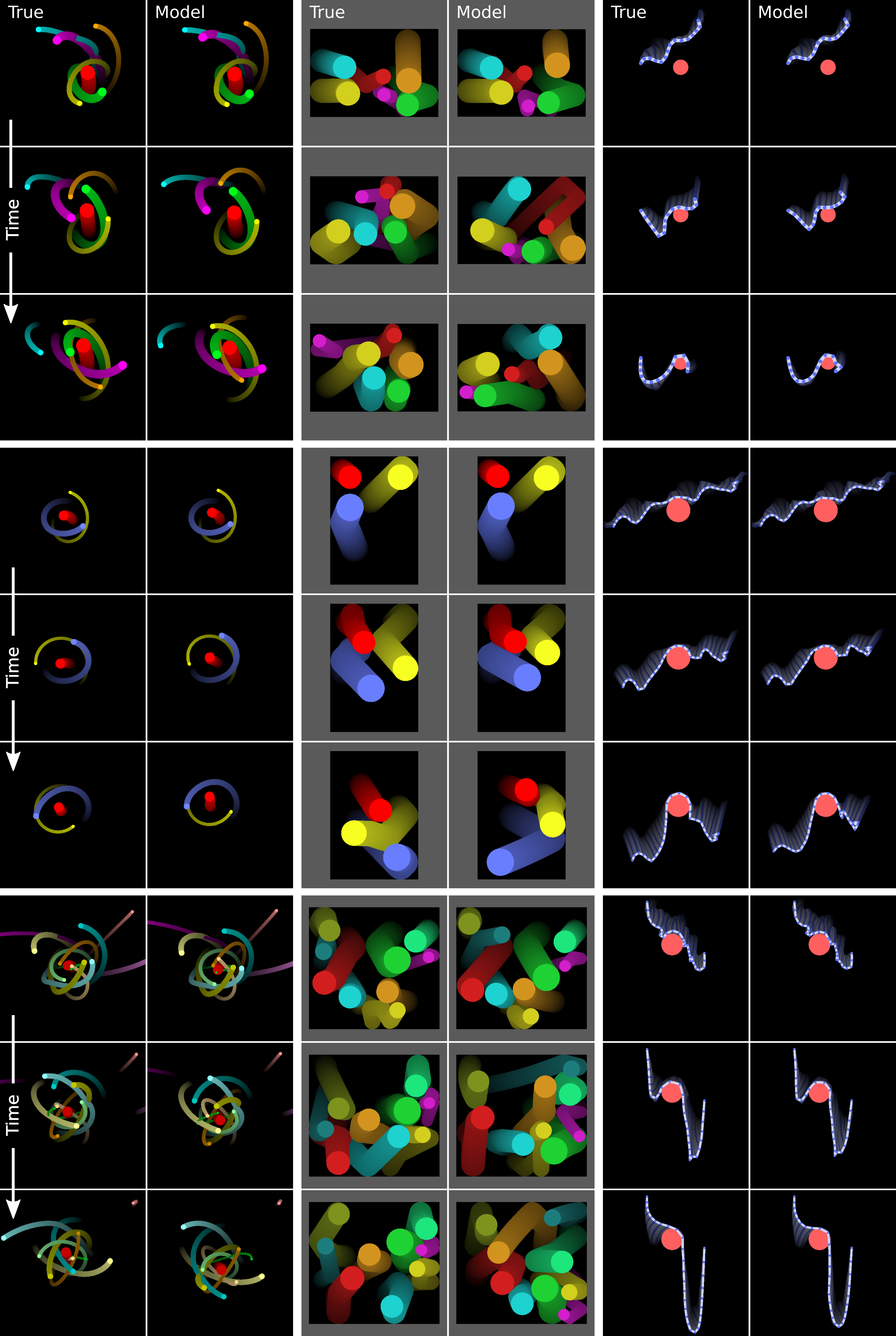}
  \caption{\small Prediction rollouts. Each column contains three
    panels of three video frames (with motion blur), each spanning
    1000 rollout steps. Columns 1-2 are ground truth and model
    predictions for n-body systems, 3-4 are bouncing balls, and 5-6
    are strings. Each model column was generated by a single model,
    trained on the underlying states of a system of the size in the
    top panel. The middle and bottom panels show its generalization to
    systems of different sizes and structure. For n-body, the training
    was on 6 bodies, and generalization was to 3 and 12 bodies. For
    balls, the training was on 6 balls, and generalization was to 3
    and 9 balls. For strings, the training was on 15 masses with 1 end
    pinned, and generalization was to 30 masses with 0 and 2 ends
    pinned. The URLs to the full videos of each rollout are in
    Table~\ref{tab:videos}.}
  \label{fig:rollouts}
\end{figure}

Our model takes as input the state of each system, $G$, decomposed
into the objects, $O$ (e.g., n-body objects, balls, walls, points
masses that represented string elements), and their physical
relations, $R$ (e.g., gravitational attraction, collisions, springs),
as well as the external effects, $X$ (e.g., gravitational
acceleration). Each object state, $o_j$, could be further divided into
a dynamic state component (e.g., position and velocity) and a static
attribute component (e.g., mass, size, shape). The relation
attributes, $R_a$, represented quantities such as the coefficient of
restitution, and spring constant. The input represented the system at
the current time. The prediction experiment's target outputs were the
velocities of the objects on the subsequent time step, and the energy
estimation experiment's targets were the potential energies of the
system on the current time step. We also generated multi-step rollouts
for the prediction experiments (Fig.~\ref{fig:rollouts}), to assess
the model's effectiveness at creating visually realistic
simulations. The output velocity, $v_t$, on time step $t$ became the
input velocity on $t+1$, and the position at $t+1$ was updated by the
predicted velocity at $t$.

\paragraph{Data}
Each of the training, validation, test data sets were generated by
simulating 2000 scenes over 1000 time steps, and randomly sampling 1
million, 200k, and 200k one-step input/target pairs,
respectively. The model was trained for 2000 epochs, randomly
shuffling the data indices between each. We used mini-batches of 100,
and balanced their data distributions so the targets had similar
per-element statistics. The performance reported in the Results was
measured on held-out test data.

We explored adding a small amount of Gaussian noise to 20\% of the
data's input positions and velocities during the initial phase of
training, which was reduced to 0\% from epochs 50 to 250. The noise
std. dev. was $0.05 \times$ the std. dev. of each element's values
across the dataset. It allowed the model to experience physically
impossible states which could not have been generated by the physics
engine, and learn to project them back to nearby, possible states. Our
error measure did not reflect clear differences with or without noise,
but rollouts from models trained with noise were slightly more
visually realistic, and static objects were less subject to drift over
many steps.

\paragraph{Model architecture}
The $f_R$ and $f_O$ MLPs contained multiple hidden layers of linear
transforms plus biases, followed by rectified linear units (ReLUs),
and an output layer that was a linear transform plus bias. The best
model architecture was selected by a grid search over layer sizes and
depths. All inputs (except $R_r$ and $R_s$) were normalized by
centering at the median and rescaling the 5th and 95th percentiles to
-1 and 1. All training objectives and test measures used mean squared
error (MSE) between the model's prediction and the ground truth
target.

All prediction experiments used the same architecture, with parameters
selected by a hyperparameter search. The $f_R$ MLP had four,
$150$-length hidden layers, and output length $\DE=50$. The $f_O$ MLP
had one, $100$-length hidden layer, and output length $\DP=2$, which
targeted the $x,y$-velocity. The $m$ and $a$ were customized so that
the model was invariant to the absolute positions of objects in the
scene. The $m$ concatenated three terms for each $b_k$: the difference
vector between the dynamic states of the receiver and sender, the
concatenated receiver and sender attribute vectors, and the relation
attribute vector. The $a$ only outputs the velocities, not the
positions, for input to $\phi_O$.

The energy estimation experiments used the IN from the prediction
experiments with an additional $\phi_A$ MLP which had one, $25$-length
hidden layer. Its $P$ inputs' columns were length $\DP=10$, and its
output length was $\DA=1$.

We optimized the parameters using Adam \cite{kingma2014adam}, with a
waterfall schedule that began with a learning rate of $0.001$ and
down-scaled the learning rate by $0.8$ each time the validation error,
estimated over a window of $40$ epochs, stopped decreasing.

Two forms of L2 regularization were explored: one applied to the
effects, $E$, and another to the model parameters. Regularizing $E$
improved generalization to different numbers of objects and reduced
drift over many rollout steps. It likely incentivizes sparser
communication between the $\phi_R$ and $\phi_O$, prompting them to
operate more independently. Regularizing the parameters generally
improved performance and reduced overfitting. Both penalty factors
were selected by a grid search.

Few competing models are available in the literature to compare our
model against, but we considered several alternatives: a constant
velocity baseline which output the input velocity; an MLP baseline,
with two $300$-length hidden layers, which took as input a flattened
vector of all of the input data; and a variant of the IN with the
$\phi_R$ component removed (the interaction effects, $E$, was set to a
$0$-matrix).

\section{Results}
\begin{figure}[t]
  \centering
    \includegraphics[width=1\textwidth]{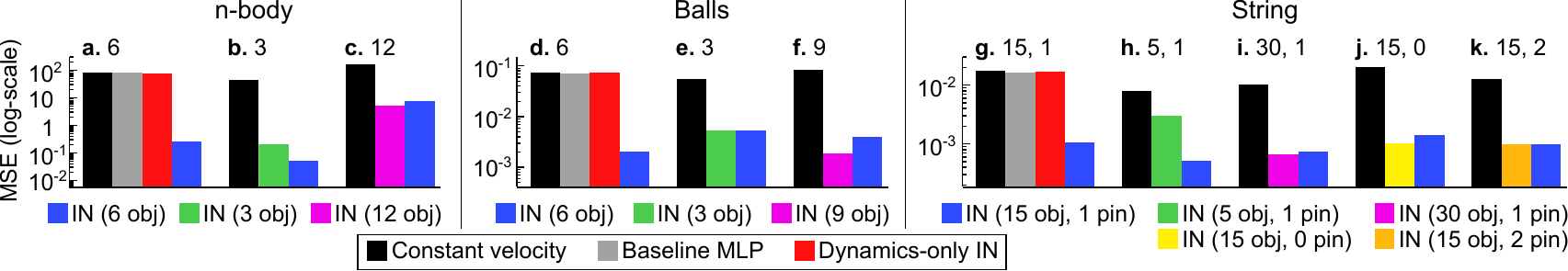}
    \caption{\small Prediction experiment accuracy and
      generalization. Each colored bar represents the MSE between a
      model's predicted velocity and the ground truth physics engine's
      (the y-axes are log-scaled). Sublots (\textbf{a-c}) show n-body
      performance, (\textbf{d-f}) show balls, and (\textbf{g-k}) show
      string. The leftmost subplots in each (\textbf{a, d, g}) for
      each domain compare the constant velocity model (black),
      baseline MLP (grey), dynamics-only IN (red), and full IN
      (blue). The other panels show the IN's generalization
      performance to different numbers and configurations of objects,
      as indicated by the subplot titles. For the string systems, the
      numbers correspond to: (the number of masses, how many ends were
      pinned).}
    \label{fig:main-bars}
\end{figure}

\paragraph{Prediction experiments}
Our results show that the IN can predict the next-step dynamics of our
task domains very accurately after training, with orders of magnitude
lower test error than the alternative models
(Fig.~\ref{fig:main-bars}a, d and g, and Table~\ref{tab:mse}). Because
the dynamics of each domain depended crucially on interactions among
objects, the IN was able to learn to exploit these relationships for
its predictions. The dynamics-only IN had no mechanism for processing
interactions, and performed similarly to the constant velocity
model. The baseline MLP's connectivity makes it possible, in
principle, for it to learn the interactions, but that would require
learning how to use the relation indices to selectively process the
interactions. It would also not benefit from sharing its learning
across relations and objects, instead being forced to approximate the
interactive dynamics in parallel for each objects.

The IN also generalized well to systems with fewer and greater numbers
of objects (Figs.~\ref{fig:main-bars}b-c, e-f and h-k, and Table~SM1
in Supp. Mat.). For each domain, we selected the best IN model from
the system size on which it was trained, and evaluated its MSE on a
different system size. When tested on smaller n-body and spring
systems from those on which it was trained, its performance actually
exceeded a model trained on the smaller system. This may be due to the
model's ability to exploit its greater experience with how objects and
relations behave, available in the more complex system.

We also found that the IN trained on single-step predictions can be
used to simulate trajectories over thousands of steps very
effectively, often tracking the ground truth closely, especially in
the n-body and string domains. When rendered into images and videos,
the model-generated trajectories are usually visually
indistinguishable from those of the ground truth physics engine
(Fig.~\ref{fig:rollouts}; see Table~\ref{tab:videos} for URLs to the
full videos of each rollout). This is not to say that given the same
initial conditions, they cohere perfectly: the dynamics are highly
nonlinear and imperceptible prediction errors by the model can rapidly
lead to large differences in the systems' states. But the incoherent
rollouts do not violate people's expectations, and might be roughly on
par with people's understanding of these domains.

\begin{table}[t]
  \caption{Rollout video URLs}
  \label{tab:videos}
  \centering
  \begin{tabular}{lll}
    \toprule
    System & True & Model \\
    \midrule
     n-body 6 & \href{https://youtu.be/otIGNTFJwpU}{https://youtu.be/otIGNTFJwpU} & \href{https://youtu.be/CXzubiwu4GM}{https://youtu.be/CXzubiwu4GM} \\
     n-body 3 & \href{https://youtu.be/zOTVzziJz24}{https://youtu.be/zOTVzziJz24} & \href{https://youtu.be/3bFrbjdTc6s}{https://youtu.be/3bFrbjdTc6s} \\
     n-body 12 & \href{https://youtu.be/JgSRaYmQyNY}{https://youtu.be/JgSRaYmQyNY} & \href{https://youtu.be/Rl44jdaIIiw}{https://youtu.be/Rl44jdaIIiw} \\
     Balls 6 & \href{https://youtu.be/9c-UPXSjOHI}{https://youtu.be/9c-UPXSjOHI} & \href{https://youtu.be/SV741g_qp8M}{https://youtu.be/SV741g\_qp8M} \\
     Balls 3 & \href{https://youtu.be/op2d9IHh3Ak}{https://youtu.be/op2d9IHh3Ak} & \href{https://youtu.be/i0lOn_v8xVg}{https://youtu.be/i0lOn\_v8xVg} \\
     Balls 9 & \href{https://youtu.be/oQoquoVJ8V0}{https://youtu.be/oQoquoVJ8V0} & \href{https://youtu.be/41y_saKvi3s}{https://youtu.be/41y\_saKvi3s} \\
     String 15, 1 pinned & \href{https://youtu.be/xiz49xOPX9c}{https://youtu.be/xiz49xOPX9c} & \href{https://youtu.be/pbk6VM_XQb4}{https://youtu.be/pbk6VM\_XQb4} \\
     String 30, 0 pinned & \href{https://youtu.be/TpixFr_ekAE}{https://youtu.be/TpixFr\_ekAE} & \href{https://youtu.be/tsE7cev0w08}{https://youtu.be/tsE7cev0w08} \\
     String 30, 2 pinned & \href{https://youtu.be/x6g1ogRG31o}{https://youtu.be/x6g1ogRG31o} & \href{https://youtu.be/ZDCZyqyrv5o}{https://youtu.be/ZDCZyqyrv5o}\\
    \bottomrule
  \end{tabular}
\end{table}

\begin{table}[t]
  \caption{Prediction experiment MSEs}
  \label{tab:mse}
  \centering
  \begin{tabular}{clccccccccccccc}
    \toprule
    Domain & Constant velocity & Baseline & Dynamics-only IN & IN \\
    \midrule
    n-body & 82 & 79 & 76 & \textbf{0.25} \\
    Balls & 0.074 & 0.072 & 0.074 & \textbf{0.0020} \\
    String & 0.018 & 0.016 & 0.017 & \textbf{0.0011} \\
    \bottomrule
  \end{tabular}
\end{table}

\paragraph{Estimating abstract properties}
We trained an abstract-estimation variant of our model to predict
potential energies in the n-body and string domains (the ball domain's
potential energies were always $0$), and found it was much more
accurate (n-body MSE $1.4$, string MSE $1.1$) than the MLP baseline
(n-body MSE $19$, string MSE $425$). The IN presumably learns the
gravitational and spring potential energy functions, applies them to
the relations in their respective domains, and combines the results.

\section{Discussion}
We introduced interaction networks as a flexible and efficient model
for explicit reasoning about objects and relations in complex
systems. Our results provide surprisingly strong evidence of their
ability to learn accurate physical simulations and generalize their
training to novel systems with different numbers and configurations of
objects and relations. They could also learn to infer abstract
properties of physical systems, such as potential energy. The
alternative models we tested performed much more poorly, with orders
of magnitude greater error. Simulation over rich mental models is
thought to be a crucial mechanism of how humans reason about physics
and other complex domains
\cite{craik1943nature,johnson1983mental,hegarty2004mechanical}, and
Battaglia et al. \cite{battaglia2013simulation} recently posited a
simulation-based ``intuitive physics engine'' model to explain human
physical scene understanding. Our interaction network implementation
is the first learnable physics engine that can scale up to real-world
problems, and is a promising template for new AI approaches to
reasoning about other physical and mechanical systems, scene
understanding, social perception, hierarchical planning, and
analogical reasoning.

In the future, it will be important to develop techniques that allow
interaction networks to handle very large systems with many
interactions, such as by culling interaction computations that will
have negligible effects. The interaction network may also serve as a
powerful model for model-predictive control inputting active control
signals as external effects -- because it is differentiable, it
naturally supports gradient-based planning. It will also be important
to prepend a perceptual front-end that can infer from objects and
relations raw observations, which can then be provided as input to an
interaction network that can reason about the underlying structure of
a scene. By adapting the interaction network into a recurrent neural
network, even more accurate long-term predictions might be possible,
though preliminary tests found little benefit beyond its
already-strong performance. By modifying the interaction network to be
a probabilistic generative model, it may also support probabilistic
inference over unknown object properties and relations.

By combining three powerful tools from the modern machine learning
toolkit -- relational reasoning over structured knowledge, simulation,
and deep learning -- interaction networks offer flexible, accurate,
and efficient learning and inference in challenging
domains. Decomposing complex systems into objects and relations, and
reasoning about them explicitly, provides for combinatorial
generalization to novel contexts, one of the most important future
challenges for AI, and a crucial step toward closing the gap between
how humans and machines think.

\small
\bibliographystyle{plain}
\bibliography{interaction-networks}

\appendix
\newpage
\section{Supplementary material}

\subsection{Experimental details}
\subsubsection{Physics engine details}
Every simulated trajectory contained the states of the objects in the
system on each frame of a sequence of 1000 one-millisecond time
steps. The parameters of each system were chosen so that a diverse set
of dynamics could unfold within the trajectory. On each step, the
physics engine took as input the current state of the system,
calculated the forces associated with each inter-entity interaction,
and applied them to the individual entities as accelerations by
dividing by the entity's mass, parameterized as inverse-mass, $a = F
m^{-1}$ in both the engine and model input. This also allows static
objects to be represented as having $m^{-1}=0$. The previous positions
and velocities, and newly computed accelerations, were input to an
Euler integrator to update the current velocities and positions of the
entities. By using a custom engine we were able to have perfect
control over the details of the simulation, and use one engine for all
physical domains. It produced trajectory rollouts that were
indistinguishable from those of an off-the-shelf simulation engine,
but was highly efficient because it could perform thousands of runs in
parallel on a GPU, allowing millions of simulated steps to be
generated each second.

\subsubsection{Physical domains}
\paragraph{n-body}
All objects in n-body systems exerted gravitational forces on each
other, which were a function of the objects' pairwise distances and
masses, giving rise to very complex dynamics that are highly sensitive
to initial conditions. Across simulation runs, the objects' masses
varied, while all other non-dynamic variables were held constant. The
gravitational constant was set to ensure that objects could move
across several hundred meters within the 1000 step rollout. The
training scenes always included 6 objects, and for testing we used 3,
6, and 12 objects. The masses were uniformly sampled from $[0.02,9]$
kg, their shapes were points, and their initial positions were
randomly sampled from all angles, with a distance in $[10,100]$ m. We
included two classes of scenes. The first, orbit systems, had one
object (the star), initialized at position $(0,0)$, with zero velocity
and a mass of $100$ kg. The planets' velocities were initialized such
that they would have stable orbits around the star, if not for their
interactions with other planets. The second, non-orbit systems,
sampled initial $x$- and $y$-velocity components from $[-3,3]$
m/s. The objects would typically begin attracting, and gave rise to
complex semi-periodic behavior.

An n-body system is a highly nonlinear (chaotic) dynamical system,
which means they are highly sensitive to initial conditions and
extended predictions of their states are not possible under even small
perturbations. The relations between the objects corresponded to
gravitational attraction. Between simulation runs, the masses of the
objects were varied, while all other non-dynamic variables were held
constant (e.g., gravitational constant) or were not meaningful (e.g.,
object scales and shapes). The gravitational force from object $i$ to
$j$ was computed as, $F_{ij}=\frac{G m_i m_j (x_i - x_j)}{\|x_i -
x_j\| ^ 3}$, where $G$ is the gravitational constant. The denominator
was clipped so that forces could not go too high as the distances
between objects went to zero. All forces applied to each object were
summed to yield the per-object total forces.

\paragraph{Bouncing balls}
The bouncing balls domain still had all-to-all object
relations--collisions--and any object could collide with any other,
including the walls. But colliding objects are more difficult to
simulate than the gravitational interactions in n-body systems, and
our bouncing balls domain also included more variability in object
attributes, such as shape, scale, and mass (as before), as well as
relation attributes, such as the coefficient of restitution. The data
distribution was much more challenging: for more than 99\% of the time
steps, a ball was not in contact with any others, and its next-step
velocity equaled its current velocity. For the remaining steps,
however, collisions caused next-step velocities that was a complex
function of its state and the state of the object it collides
with. The training scene contained 6 balls inside a box with 4 walls,
and test scenes contained either 3, 6, or 9 balls. The balls' radii
were sampled uniformly from $[0.1,0.3]$ m, and masses from
$[0.75,1.25]$ kg. The walls were static rectangular boxes, positioned
so that the horizontal and vertical lengths of area they enclosed
varied independently between $[1,3]$ m. The balls' initial positions
were randomly sampled so that all balls fit within the box and did not
interpenetrate any other object, with initial $x$- and $y$-velocity
components sampled uniformly from $[-5,5]$ m/s. The restitutions, an
attribute of their collision relations, was sampled uniformly from
$[0.4,1]$.

Rigid body collision systems are highly nonlinear (chaotic) dynamical
systems, as well. Collision forces were applied between objects when
they began to interpenetrate, using two-step process: collision
detection between all objects, and resolution of detected
collisions. A detected collision between two objects meant that their
shapes overlapped and their velocities were causing them to approach
each other. To resolve a partially inelastic collision, the
post-collision velocities of each object were computed, and forces
appropriate to effect these velocities on the subsequent time step
were then calculated and applied the the objects. This resulted in
realistic bouncing ball trajectories.

\paragraph{String}
Our string domain used multiple types of relations (springs and
collisions), relation structures that were more sparse and specific
than all-to-all, and variable external effects. Each scene contained
one string, comprised of point masses connected by springs, and one
static, rigid circle positioned below the string. Gravitational
acceleration, varied across simulation runs, was applied so that the
string always fell, usually colliding with the static object. Each
training scene contained a string with 15 point masses, and test
scenes contained either 5, 15, or 30 mass strings. In training, one of
the point masses at the end of the string, chosen at random, was
always held static, as if pinned to the wall, while the other masses
were free to move. In the test conditions, we also included strings
that had both ends pinned, and no ends pinned, to evaluate
generalization. The string's masses were sampled from $[0.05,0.15]$
kg. The springs' rest lengths were set to $0.2$ m, spring constants to
$100$, and damping factors to $0.001$. The static object's
$x$-position was sampled from $[-0.5,0.5]$ m, $y$-position was sampled
from $[-1,-0.5]$ m, and radius from $[0.2,0.4]$ m. The string
mass-rigid object coefficient of restitution was sampled from
$[0,1]$. The gravitational acceleration applied to all non-static
objects was sampled from $[-30,-5]$ m/s$^2$.

The spring force from object $i$ to $j$ was computed using Hooke's law
as, $F = C_{s} (1 - \frac{L}{\|x_i - x_j\|}) (x_i - x_j)$, where
$C_{s}$ is the spring constant and $L$ is the spring's rest length. A
damping factor, which was proportional to the difference in the
objects' velocities, was also applied. Collision forces between the
string's masses and the static, rigid object were calculated as
described in the bouncing balls section above.

\subsection{Results details}
\begin{table}[t]
  \caption{Prediction experiments - Generalization MSEs}
  \label{tab:mse-gen}
  \centering
  \begin{tabular}{lccccccccccccc}
    \toprule
    & \multicolumn{2}{c}{n-body} & \multicolumn{2}{c}{Balls} & \multicolumn{4}{c}{String} \\
    \cmidrule(lr){2-3}
    \cmidrule(lr){4-5}
    \cmidrule(lr){6-9}
    Model & 3 & 12 & 3  & 9 & 5, 1 & 30, 1 & 15, 0 & 15, 2 \\
    \midrule
    Const. vel. & 45 & 166 & 0.054 & 0.084 & 0.0080 & 0.010 & 0.020 & 0.012 \\
    IN (within) & 0.21 & \textbf{5.2} & \textbf{0.0053} & \textbf{0.0019} & 0.0030 & \textbf{0.00066} & \textbf{0.0010} & \textbf{0.00097} \\
    IN (transfer) & \textbf{0.052} & 7.8 & \textbf{0.0053} & 0.004 & \textbf{0.00052} & 0.00074 & 0.0014 & \textbf{0.00097} \\
    \bottomrule
  \end{tabular}
\end{table}
The prediction experiment's generalization performance MSEs are shown
Table~\ref{tab:mse-gen}.
The URL links for the videos associated with all still frames in
Figure~\ref{fig:rollouts} are in Table~\ref{tab:videos}.

\end{document}